\documentclass[11pt,a4paper]{article}
\usepackage{times}
\usepackage{graphicx}
\usepackage{amssymb}
\usepackage{url,hyperref}

\usepackage[hyperref]{acl2021}

\usepackage{latexsym}

\usepackage{soul}
\usepackage{graphicx}
\usepackage{microtype}

\aclfinalcopy 

\title{Countering Online Hate Speech: An NLP Perspective}

\author{
Mudit Chaudhary, Chandni Saxena, Helen Meng \\
The Chinese University of Hong Kong \\
\texttt{muditchaudhary@cuhk.edu.hk}, \texttt{csaxena@cse.cuhk.edu.hk}\\ \texttt{hmmeng@se.cuhk.edu.hk}
}

\date{}

\begin{document}
\maketitle
\begin{abstract}
Online hate speech has caught everyone’s attention from the news related to the COVID-19 pandemic, US elections, and worldwide protests. Online toxicity — an umbrella term for online hateful behavior, manifests itself in forms such as online hate speech. Hate speech is a deliberate attack directed towards an individual or a group motivated by the targeted entity’s identity or opinions.  The rising mass communication through social media further exacerbates the harmful consequences of online hate speech.  While there has been significant research on hate-speech identification using Natural Language Processing (NLP), the work on utilizing NLP for prevention and intervention of online hate speech lacks relatively. This paper presents a holistic conceptual framework on hate-speech NLP countering methods along with a thorough survey on the current progress of NLP for countering online hate speech. It classifies the countering techniques based on their time of action, and identifies potential future research areas on this topic.

\end{abstract}
\section{Introduction}
Internet is a double-edged sword. While it offers many benefits, it has also become a breeding ground for radicalism, violent ideas and hate speech. Various studies have been conducted to analyze extent of online hate speech. A cross-sectional study by \cite{Keipi} showed that 42\% of 15 to 30 year-olds from United Kingdom, United States, Finland, and Germany experienced hateful materials on social media. 

Online hate speech took center stage in 2020 with the COVID-19 pandemic, worldwide protests, anti-Asian rhetoric, race-related violence, and the US presidential elections. Hate speech has many negative effects for citizens and creates societal tensions \cite{ziems2020racism, league2016anti}. These negative effects were exacerbated by social isolation \cite{Silva2020}. A rise in hate-crimes and online hate speech was also reported during the 2020 US elections\footnote{www.sandiegouniontribune.com/news/public-safety/story/2020-10-31/hate-crimes-surge-presidential-elections}\footnote{ www.fortune.com/2020/12/23/linkedin-content-moderation-hate-speech-misinformation-2020-election-coronavirus/}. In the first quarter of 2020, a popular social media platform removed 10 million posts for violating its hate speech policy\footnote{www.fortune.com/2020/05/12/facebook-removes-10-million-hate-speech-posts/}. 

\cite{Salminen2020} define online toxicity as `hateful communication that is likely to cause an individual user leave a discussion.' Online toxicity is a broader phenomenon that can manifest itself in various forms such as hate speech, cyberbullying, trolling, abuse, etc. \cite{salminen2020developing}. In this paper, we focus on hate speech which has significant similarities with other forms of online toxicity. The definition of `hate speech' is very subjective and its interpretation is contingent upon the cultural, economical and racial contexts. The many definitions of hate speech shows a general lack of consensus, which makes the task of identifying and countering it challenging because it invites differing legal, ethical and socio-technical perspectives \cite{macavaney2019hate}. \cite{10.1145/3232676} projected the definitions of hate speech from different sources into four dimensions -- (i) hate speech is to incite violence or hate, (ii) hate speech is to attack or diminish, (iii) hate speech has specific targets and (iv) humor has a specific status. In addition, they observe that the dimension `hate speech has specific target' and `hate speech is to attack or diminish' is common among many definitions. Referencing previous work, this paper adopts a working definition for hate speech as follows -- A negative form of communication that deliberately attacks an individual or a group motivated by the targeted entity's identity, characteristics, or opinions; and aims to humiliate, demotivate, harass, or incite violence.

Moderating online hate speech, referred to as `online content moderation,' has resulted in worsening mental health of the content moderators due to graphic content, descriptions and hateful posts. It has been reported that moderators suffer from Post-Traumatic Stress Disorder (PTSD) as a consequence of their work\footnote{https://www.bbc.com/news/technology-52642633}. Countering online hate speech through Natural Language Processing (NLP) has emerged as an important field of research due to its potential of automating the process at scale and also reducing the workload and mental stress on the moderators. NLP has shown great potential in tasks such as automated text classification, natural language generation and content style-transfer \cite{devlin2019bert, radford2019language, shen2017style}. Consequently, it is being used to tackle online hate speech. There has been development of various benchmarking datasets for NLP-based online hate speech detection and intervention models \cite{fortuna-etal-2020-toxic, mathew2020hatexplain, DBLP:journals/corr/abs-1909-04251}. 

Countering hate speech includes detection, prevention and intervention methods to control online hate speech. A significant amount of research is being done on using NLP for hate speech detection~\cite{schmidt2017survey}, but there is a paucity of work on NLP-based prevention and intervention methods. Hence, in this paper we focus more on prevention and intervention methods instead of detection methods. This theme paper aims to provide comprehensive overview of the latest techniques inside a holistic conceptual framework to counter online hate speech. We classify the methods into the \emph{proactive} and \emph{reactive} categories -- each has different pros and cons as well as ethical concerns. We intend for this extensive survey and the framework to help identify promising directions for future research. 

\section{Definitions}
This section provides the definitions used in the proposed conceptual framework. 

\noindent\textbf{Roles.} We identify three roles that are relevant for the study:
\begin{itemize}
    \item \textbf{Author (First-person):} The author of the online content or interaction. 
    \item \textbf{Moderator (Second-person):} The moderator refers to the internet platform's internal systems. It can either be a human moderator, automated moderator or both.
    \item \textbf{Consumer (Third-person):} The consumer is someone who can see the released content and is not a moderator. The consumer can come from the general public or have a connection with the author. The connection with the author is platform-specific, e.g., Friends on Facebook, Follower on Instagram.
\end{itemize}

\noindent\textbf{Countering}: Countering broadly refers to ways that restrict online hate speech. Countering methods can include:
\begin{itemize}
    \item Detection of released hate speech followed by intervention methods; and/or
    \item Prevention of unreleased hate speech by potential hate speech prediction methods, followed by some intervention methods
\end{itemize}

\noindent\textbf{Reactive Countering Methods.} Reactive methods work retrospectively by using detection methods to detect hate speech in previously posted content followed by some intervention strategy. 
\begin{itemize}
    \item \textbf{Act on}
    \begin{itemize}
        \item Content visible to the author(s), moderator(s), and consumer(s); and/or
        \item Author(s)
    \end{itemize}
    \item \textbf{Act when}
    \begin{itemize}
        \item Content released to consumer(s)
    \end{itemize}
    \item \textbf{Act to}
    \begin{itemize}
    \item  Curb hate speech content that has already been released from causing further harm
    \end{itemize}
    
    \noindent\textbf{Proactive Countering Methods.} Proactive methods are preventative and try to intervene before the hate speech content reaches the consumers.
\end{itemize}
\begin{itemize}
    \item \textbf{Act on}
    \begin{itemize}
        \item Content visible to the author(s) and moderator(s), or no one (content not created); and/or
        \item Author(s)
    \end{itemize}
    \item \textbf{Act when}
    \begin{itemize}
        \item Content is not yet released to the consumer(s); and/or
        \item Hate speech is not yet created, e.g., when the method predicts that a certain conversation or user will lead to creation of hate speech in the future
    \end{itemize}
    \item \textbf{Act to}
    \begin{itemize}
    \item  prevent potential hate speech content from reaching the consumers
    \end{itemize}
\end{itemize}

\section{Framework}
\begin{figure*}

\centering
\includegraphics[width=\linewidth]{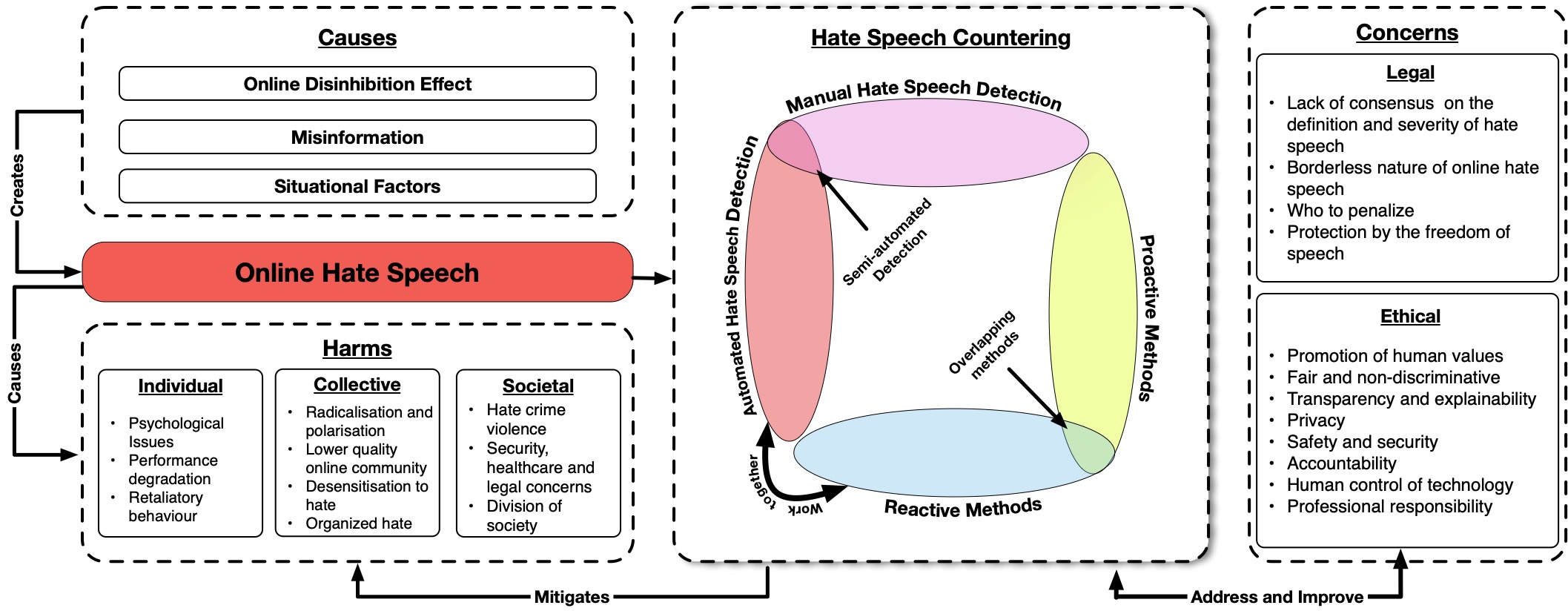}
\caption{Countering online hate speech conceptual framework. Ethical concerns from \cite{kiritchenko2020confronting}.}
\label{Figure:1}
\end{figure*}

We propose a conceptual framework for countering online hate speech as shown in Figure~\ref{Figure:1}. 
The framework includes the major components pertinent to online hate speech countering research: \emph{causes}, \emph{consequences}, \emph{countering methods}, and \emph{concerns}. 

We design and present this framework with three goals in mind: (1)~provide an overview of latest and major research in all components, (2)~encourage the user to consider the impact of different components on the overall system, and (3)~help identify future directions of research. 

We will later discuss how different components relate to each other and how it can help identify future research directions.

\subsection{Causes of Hate speech}
Online hate speech is not just a technological issue but also a social issue. Searching for the roots of online hate speech leads to an non-exhaustive list. In this section, we provide a brief overview of the major factors that lead to hate speech. We take into consideration the online disinhibition effect to understand the innate causes \cite{doi:10.1089/1094931041291295}, the work of \cite{10.1145/2998181.2998213} for situational causes, and some other studies.
\\\\
\noindent{\textbf{Online disinhibition effect.}}
This effect emerges when people are online and feel less restraint regarding their actions, and may act more intensely or frequently than they would offline, especially anonymously \cite{doi:10.1089/1094931041291295}. Two types of disinhibitions are -- benign and toxic; where toxic disinhibition leads to hateful, deviant or extreme behaviour online \cite{doi:10.1089/1094931041291295}. \cite{lai2016cyberbullying} also found a positive correlation between the \cite{udris2014cyberbullying}'s Online Disinhibition Scale (ODS)  and cyberbullying.
\\\\
\noindent{\textbf{Misinformation.}}
\cite{DelVicario554} performed a large-scale quantitative analysis of the diffusion of fake news and conspiracy theories on Facebook. In their study, they observe that the users tend to select the content that goes along with their narrative and ignore other news, leading to the formation of `echo-chambers.' When the primary drivers of these echo-chambers is fake news, it can cause polarization, mistrust, rumors, and paranoia \cite{DelVicario554}. This polarization and paranoia often manifests itself as hate speech when it is targeted towards an individual or a group. An example of this is the recent Anti-asian rhetoric and online hate speech due to COVID-19 \cite{ziems2020racism}.
\\\\
\noindent{\textbf{Situational factors.}}
\cite{10.1145/2998181.2998213} observe certain situational factors that cause online hate speech: personal mood, discussion context, and contagiousness of hate speech. Personal dissatisfaction, bad mood and anger increases aggression towards others which can lead to malicious behaviour online \cite{10.1145/2739042, 10.1145/2998181.2998213}. According to their study, the immediate context of the discussion can mold the direction of conversation. Moreover, the authors observe that a single malicious user or post can lead to multiple users engage in hate speech proliferation.

\subsection{Consequences of Hate Speech}
Various studies support the dire consequences emerging from the prevalence of hate speech. It not only gives rise to crimes but also far-fetching psychological problems. In this section, we refine the harms loosely categorised in \cite{salminen2020developing}'s framework. The harms are divided into three levels: \emph{Individual}, \emph{Collective}, and \emph{Societal}.
\subsubsection{Individual Harms}
Individual harms refer to the consequences acting at an individual level. Hate speech leads to several negative psychological effects on individuals such as depression, anxiety, drug abuse, insomnia, and in some extreme cases incite suicidal thoughts \cite{lai2016cyberbullying, pmid17954938, pmid20658375}. It can also lead to performance degradation at work or school, promote self-harm and retaliatory behaviour \cite{lai2016cyberbullying}.

\subsubsection{Collective Harms}
Collective harms are caused in a group of individuals. On a collective level it can lead to radicalization and polarization \cite{doi:10.1177/0002764202045006003, DelVicario2016}, lower quality of the online community \cite{Kumar_2018}, and insecurity \cite{CHETTY2018108}. It can also cause desensitisation to hateful statements in online users \cite{https://doi.org/10.1002/ab.21737} and lead to organized hate-speech against marginalized groups \cite{doi:10.1080/00131857.2020.1802818}.

\subsubsection{Societal Harms}
Societal harms deal with the harms caused to a collection of societal groups i.e. society. The creation of increasingly polarized online groups divides the society due to the collective trauma experienced by the targeted group during online browsing. In extreme cases, it can also result in hate crime violence \cite{ecri2016general, bakalis2015cyberhate} which adds to security, healthcare and legal concerns. 

\subsection{Hate speech countering methods}
Taking inspiration from the work of \cite{jurgens-etal-2019-just}, we classify the countering methods into two categories: \emph{proactive} and \emph{reactive}. In Section~2, we develop definitions of \emph{reactive} and \emph{proactive} to develop further from previous works. We acknowledge that some methods can belong to both categories and their classification depends on the manner in which they are applied.

\subsubsection{Reactive}
As per the definition, reactive methods deal with hate speech events after they have occured. We broadly identify three types of reactive strategies in which Natural Language Processing can help curb online hate speech: \emph{counter-speech generation}, \emph{style transfer}, and \emph{automated and semi-automated moderation}. It is to be noted that reactive methods work in tandem with hate speech detection methods. Hence, we will also briefly discuss selected works of hate speech detection because they facilitate human intervention.
\\\\
\noindent\textbf{Counter-speech generation.}
Counter-speech generation is a relatively new strategy to curb online hate speech. \cite{garland2020countering} define counter-speech as \emph{ ``a generated response to online hate in order to stop and prevent the spread of hate speech, and if possible change perpetrators’ attitudes about their victims.''} Counter-speech is gaining interest of online moderators as it does not infringe upon people's freedom of expression, making it a good alternative to deletion and blocking \cite{gagliardone2015countering, benesch2014countering, silverman2016impact}. \cite{wright-etal-2017-vectors} explore different relationship dynamics of counter-speeches on Twitter and identify the potential of deep learning in automated classification and generation.

One of the challenges for performing NLP based counter-speech generation or classification is the lack of annotated datasets. To resolve this, various datasets have been recently curated. CONAN is a multilingual dataset of counter-speech \cite{chung-etal-2019-conan}. It consists of 4,078 hate speech/counter-speech pairs with the following annotations -- presentation of facts, pointing out hypocrisy or contradiction, warning of consequences, affiliation, positive tone, negative tone, humor, counter-questions, and other. It also adds sub-topic annotations for the pairs, e.g., culture, crimes, terrorism, etc. \cite{DBLP:journals/corr/abs-1909-04251} released two benchmark datasets. They collected hate speech conversations from Gab and Reddit. The counter-speech to the hate speech content was then created by Amazon Mechanical Turk workers. \cite{tekiroglu2020generating} focused on providing a framework for generating quality data for counter-speech generation. Moreover, they use the GPT-2 model to generate counter-speech silver data followed by filtering and expert validation/editing. 

If counter-speech can be detected as a response to hate speech, the platforms can increase their visibility to other consumers so as to change the ongoing hateful narrative. \cite{mathew2019thou} proposed an ensemble method (SVM with XGBoost + TF-IDF + Bag-of-Words) achieving 0.715 F1-Score on their own curated counter-speech dataset from YouTube comments. Another ensemble based approach by \cite{garland2020countering} achieved 0.76-0.97 accuracy on their dataset consisting of tweets.

Due to the availability of new counter-speech datasets and crowdsourcing resources, work on automated counter-speech generation has shown a significant uptick. \cite{pranesh2021towards} used the dataset from \cite{DBLP:journals/corr/abs-1909-04251} to train BART\cite{lewis-etal-2020-bart}, BERT \cite{devlin2019bert} and DialoGPT \cite{zhang2020dialogpt} models for an automated counter-speech generation. ParityBOT takes a slightly different approach by using curated counter-speeches on Twitter \cite{cuthbertson2019women}. ParityBOT automatically counters abusive tweets targeted towards women in politics by sending curated counter-speeches that support female leaders.
\\\\
\noindent\textbf{Neural Style Transfer.}
Neural style transfer is another alternative to traditional moderation strategies. Using neural style transfer, we can automatically neutralize parts of hate-speech content to make it more polite without impinging upon one's freedom of speech. \cite{carton-etal-2018-extractive} use extractive adversarial networks to show parts of speech that are hateful. Using the extracted suspicious parts from the extractive adversarial network and other methods to extract suspicious parts \cite{pavlopoulos-etal-2017-deeper, svec-etal-2018-improving, noever2018machine}, we can use style transfer \cite{prabhumoye2018style, yang2019unsupervised, nogueira-dos-santos-etal-2018-fighting, sennrich-etal-2016-controlling} to either modify them to their more acceptable counterparts or redact them. 
\\\\
\noindent\textbf{Automated and Semi-automated moderation.}
Online hate speech moderation by human moderators is one of the traditional methods of content moderation. The human moderators assess the flagged content and decide the action -- Delete, Suspend or Ignore. Given the scale of social media, it becomes impossible for a limited number of human moderators to moderate all the flagged comments. According to a study conducted on 63M Wikipedia talk page comments, only an estimated 17.9\% of the attacking comments were moderated within seven days \cite{wulczyn2017ex}. Smaller platforms or non-profit platforms do not have the resources to employ human moderators. Furthermore, the mental stress caused by the exposure of hateful behavior \cite{iyer-barve-2020} on the moderators calls for automated or semi-automated moderation. 

\cite{pavlopoulos-etal-2017-deeper} proposed a Recurrent Neural Network (RNN) operation on word embeddings to assist online moderation. They develop a classification specific attention mechanism that improves hate speech detection and also highlights suspicious words. The highlighted words help the moderators make their decision more efficiently. In their semi-automated approach, the comments with high confidence are automatically regulated while the ones with low confidence are sent to the moderators to decide. Following the same rationale, \cite{svec-etal-2018-improving} developed a two-step method which consists of a hate speech classifier and suspicious word detector. They use RCNN recurrent cells \cite{lei-etal-2016-semi} in their classifier. The suspicious word detector uses a reinforcement learning based approach \cite{lei-etal-2016-rationalizing}.

To make the hate speech classification decision more interpretable for the moderators, \cite{risch2018delete} created a non-deep learning based logistic regression model which predicts the probability of appropriateness for a comment. The model inputs 3 types of features: (1)~Comment features, (2)~User features, and (3)~Article features. According to the authors, their implementation can give the moderators insight on the influence of different features on the classification decision. Another approach tries to improve the detection of hate speech by adding user embeddings or user type embeddings to an RNN model \cite{pavlopoulos-etal-2017-improved}. The user embeddings are based on the amount of rejections in the past. The user embeddings add an additional context to aid the detection model, which can potentially assist the moderators in making the decision for low confidence detections. 

For automated moderation, in addition to deletion or suspension of the content, the system can use the alternative strategies mentioned before such as neural style transfer and counter-speech generation. The final action of automated moderation system is a system design concern that can be decided by the platform based on their context.
 
 \subsubsection{Proactive}
As stated in the definition, proactive methods deal with hate speech before a hate event occurs so that the hate speech content is not released to the consumers. These methods provide a timely note to the moderator to intervene before it evolves into severe hate speech. Proactive methods are prone to ethical concerns as discussed in the later section of this paper.  Based on NLP interventions, we identify `preemptive moderation' as a broad class of proactive strategies. It is worth noting that \emph{potential} hate speech prediction methods (not to be confused with \emph{``hate speech detection''}) act as tools for assisting proactive methods. In this section, we also come up with suggestions on using other NLP interventions that could work with predictive detection of hate in proactive settings for future research. 
 \\\\
\noindent\textbf{Preemptive Moderation.} This is a proactive step to respond to hate speech by preventing individuals' hate content from expanding in severity. Recent works have shown that it is possible to predict if a conversation will turn into a hate speech
in the future ~\cite{brassard2020using,karan2019preemptive,chang2019trouble,zhang2018conversations}. These predictive models can prevent harm by employing intervention steps early-on. \cite{karan2019preemptive} explored potential hate speech detection and worked on preemptive moderation. The authors employed a linear SVM model using TFIDF-weighted unigrams and bigrams as well as a BiLSTM model using the word embeddings of Wikipedia conversations data. \cite{zhang2018conversations} introduced a model to detect early warning signs of toxicity and used linguistic features for intervention strategies. The authors investigated a moderation framework for capturing the relation between politeness strategies and the conversation's future trajectory. Working on the presence and intensity prediction on Instagram comment threads, \cite{liu2018forecasting} predicted hostility presence and hostility intensity in Instagram comments. They used multiple features, ranging from n-grams and word vectors to user activity, trained a logistic regression model with L2-regularization, and examined the impact of sentiment on forecasting tasks. The authors suggested to use the model for preemptive prediction on prior hostility features and employ intervention techniques such as comment filtering and comment controls. \cite{brassard2020using} encoded conversations features (text features and sentiment features) of the messages for preemptive prediction of hate speech.
The authors validated the generality of results and used a model for predicting hate speech on gaming chat data.

In their positional paper, \cite{jurgens2019just} argue for
a need to boost research for developing proactive strategies for hate speech intervention that provides assistance to authors, moderators, and social platforms to curb the hate phenomenon before it occurs. Based on the same rationale, we suggest some future directions: (1)~One can employ technology to quarantine hate speech in a proactive course of action and further use appropriate intervention strategy to curb the effect of hate before it is released to the consumers. Quarantining in this context means temporarily holding off content from consumers for review. This framework also provides a balance between freedom of expression and relevant censorship. (2)~Existing researches show progress in the direction of potential hate detection in proactive state and use of more comprehensive features including history data (past comments)~\cite{chelmis2019minority}, text and tree structure of past communication~\cite{hessel2019something}, and behavioral sequence with n-gram features~\cite{tshimula2020predicting}. These models can be used to support proactive approaches and offer help for countering hate speech. 

\subsection{Concerns}
In this section, we briefly discuss ethical and legal concerns surrounding online hate speech countering. The ethical concerns can be addressed by the implementations at system design-level through socio-technical contemplation and intervention. However, addressing legal concerns often need a policy-level change making it more challenging as it precedes a lengthy process requiring multilateral support.
\subsubsection{Ethical}
Countering online hate speech brings various ethical concerns that need to be addressed to develop fair and reliable countering systems. One of the most prominent open-ended ethical dilemma is how to counter hate speech without infringing one's freedom of speech. We observe counter-arguments from both sides. A more authoritarian approach may argue that blocking hate speech is not concerned with freedom of speech as it is a threat to public safety, while a more liberal approach may argue that hate speech should be countered with more speech instead of censorship, which makes counter-speech and neural style transfer become attractive fields of research.

\cite{kiritchenko2020confronting} present a detailed and comprehensive analysis of countering online hate speech from an ethical perspective. They identify eight ethics theme that we need to consider while designing the countering methods. We briefly discuss these eight themes in form system design questions:
\begin{itemize}
    \item \textbf{Does the system promote human values?} E.g., infringing freedom of speech, equality and dignity of the citizens.
    \item \textbf{Is the system fair and non-discriminative?} E.g., fixing a system that unfairly treats content based on author's demographics or identity.
    \item \textbf{Is the system transparent and explainable?} E.g., explaining why a certain post was considered hateful and why the corresponding action was taken to curb its influence.
    \item \textbf{Does the system respect user's privacy?} E.g., avoiding model training on user's personal data without consent or using privacy-respecting learning methods.
    \item \textbf{Does the system ensure safety and security of the users?} E.g., considering the consequences of wrong predictions such as ignoring content that incites offline violence.
    \item \textbf{Does the system ensure accountability of its decisions?} E.g., auditing system design and creating appeal procedures.
    \item \textbf{Does the system allow human to take control?} E.g., manually overriding automated wrong decisions.
    \item \textbf{Do the system operators ensure professional responsibility?} E.g., not using the systems for deliberate unfair practices such as target group surveillance.
\end{itemize}

We can find systems that try to address some of the concerns, such as mitigating biases \cite{sap-etal-2019-risk, xia-etal-2020-demoting, zueva2020reducing}, developing interpretable and semi-automated moderation approaches \cite{pavlopoulos-etal-2017-deeper, svec-etal-2018-improving, risch2018delete}.

\subsubsection{Legal}
Legal concerns deal with policy-level interventions which are not in control of the system designers and require legislative actions to resolve. We identify the following legal concerns pertaining to countering online hate speech:

\begin{itemize}
    \item \textbf{Lack of consensus on definition and severity of hate speech:} As mentioned earlier, there is no clear definition of hate speech, which makes it harder to legislate laws for it. Another issue that can be raised is when an instance of hate speech may become liable to legal action. This point may be elusive because the severity of hate speech is highly subjective to the contexts in which it happens and often the consequences of the hate speech are intangible in nature such as psychological effects \cite{mcgonagle2013council}.
    
    \item \textbf{Borderless nature of the Internet and multiple justice systems:} The Internet is borderless and international in nature and so is the reach of online hate speech. This makes the legislation of online hate speech challenging \cite{banks2010regulating}. This issue calls for more multilateral support between different governments and justice systems to curb severe online hate speech. Towards the direction of multilateral support, the Council of Europe introduced a protocol to address online hate speech and invited other countries to adopt it \cite{council2003additional, council_of_europe_2016}.
    
    \item \textbf{Who to penalize: } Another topic of debate is who to penalize for the online hate speech. Often the hate speech is perpetuated using anonymous accounts, which makes it hard for the enforcement agencies to track down the perpetrators \cite{gagliardone2015countering}. There have also been calls to penalize the platforms instead of the perpetrators because they failed to moderate hate speeches\footnote{https://www.reuters.com/article/us-germany-fakenews-idUSKBN16L14G}.
    
    \item \textbf{Protected by the Freedom of Speech: } Another issue that was previously mentioned is the protection of hate speech by laws \cite{citron2011intermediaries}. Speech, hateful or not, is nevertheless a form of expression and silencing any speech can infringe upon one's freedom of speech. It presents a serious challenge to legislators on how to balance the freedom of speech with security and equality. Recently, some countries have introduced rules to restrict hate speech based on their content and likelihood to cause harm \cite{gagliardone2015countering}.
\end{itemize}

\section {Discussion}
In this section, we discuss some important parts of the framework and how we can use the framework to assist in identification of future research directions.

The rationale behind differentiating between the countering methods between \emph{proactive} and \emph{reactive} is because they entail different concerns and harms. \emph{Reactive} methods are less likely to invoke the argument of infringing the freedom of speech, as there is clear evidence of the hate speech event. In contrast, \emph{proactive} methods act before the hate speech event happens, putting them in an ethical gray area as the evidence of future hate speech can be unclear and circumstantial. Some \emph{proactive} methods require close surveillance of the conversations as they develop, which increases the computational overhead and makes them prone to surveillance and privacy infringement issues. Moreover, as \emph{proactive} methods try to forecast the future, the effect of biases can be much more profound based on the user's demographics, identity, and network connections. \emph{Proactive} methods can better limit harm as hate speech is not yet visible to the consumers, whereas \emph{reactive} methods need to be concerned about the amount and severity of harm, which depend on the time delay between posting and intervention. Both types of methods come with their pros and cons, which provides a potential future direction for balancing these methods for more beneficial outcomes.

The framework states various causes, ethical and legal concerns. We encourage the readers to mix and match  different methodologies with the concerns to improve the countering systems. For example, developing a \emph{pre-emptive moderation system} that respects user's privacy using privacy respecting methods such as federated learning \cite{10.1145/3298981}. Moreover, the causes of hate speech can inform development of \emph{proactive} methods of countering online hate speech. The framework also provides a checklist of concerns to be addressed in future research and development.

\section {Conclusion}
This paper provides a comprehensive conceptual framework for research on countering online hate speech, including an extensive survey of the latest methods and pertinent studies. It identifies the ethical concerns faced by \emph{proactive} and \emph{reactive} methods. It also discusses the ethical concerns checklist inspired by \cite{kiritchenko2020confronting} to help the readers develop more reliable and ethically robust countering systems.

We recognize that NLP is just one of the many tools that can help curb online hate speech. A complete solution will require multilateral support among researchers, policymakers, and the average citizen, with a focus on ethical considerations.



\end{document}